\documentclass{article}

\usepackage{booktabs}
\usepackage{PRIMEarxiv}
\usepackage{tcolorbox}
\tcbuselibrary{breakable}
\usepackage[utf8]{inputenc} 
\usepackage[T1]{fontenc}    
\usepackage{url}            
\usepackage{booktabs}       
\usepackage{amsfonts}       
\usepackage{nicefrac}       
\usepackage{microtype}      
\usepackage{lipsum}
\usepackage{fancyhdr}       
\usepackage{graphicx}       
\usepackage{xcolor}         
\usepackage{array}          
\usepackage{amsmath}        
\usepackage{tikz}
\usetikzlibrary{positioning, shapes, arrows, fit, backgrounds, calc}
\usepackage{xcolor}
\usepackage{float}
\usepackage{enumitem}       
\usepackage[minnames=6,maxnames=6,sorting=none,backend=bibtex]{biblatex}
\usepackage{hyperref}       
\usepackage{makecell}       
\usepackage{fancyvrb}
\usepackage{listings}

\graphicspath{{media/}}     

\title{Bench to the Future:\\A Pastcasting Benchmark for Forecasting Agents

}

\author{
  \And
  {\hspace{17em}}
  FutureSearch \\
  \And
  {\hspace{17em}}
  \And
  {\hspace{6em}}
  \And
  Jack Wildman \\
  \And
  Nikos I. Bosse\\
  \And
  Daniel Hnyk\\
  \And
  Peter M\"{u}hlbacher\\
  \And
  {\hspace{6em}}
  \And
  {\hspace{5em}}
  \And
  Finn Hambly\\
  \And
  Jon Evans\\
  \And
  Dan Schwarz\\
  \And
  Lawrence Phillips\\
  \And
  {\hspace{5em}}
}

\begin{document}
\maketitle

\begin{abstract}

  Forecasting is a challenging task that offers a clearly measurable way to study AI systems. Forecasting requires a large amount of research on the internet, and evaluations require time for events to happen, making the development of forecasting benchmarks challenging. To date, no forecasting benchmark provides a realistic, hermetic, and repeatable environment for LLM forecasters. We introduce Bench To the Future (BTF), a "pastcasting" benchmark with hundreds of high-quality questions for which the resolution is already known. Each question is accompanied by a large offline corpus of tens of thousands of relevant web pages, enabling a way to elicit realistic "forecasts" on past events from LLMs. Results suggest that our pastcasting environment can produce results comparable to those based on forecasts using the internet on at-the-time unresolved questions. We show results benchmarking agent and chain-of-thought forecasting approaches using several LLMs, including the recently-released Claude 4 models, and demonstrate BTF's ability to track steady forecasting capability progress over time. We intend this to be a living benchmark, with new questions added continually to account for increasing training data cutoff dates. We invite researchers to contact us at \url{hello@futuresearch.ai} to utilize our benchmark or tooling for their own research.

\end{abstract}

\section{Introduction}

Agents based on Large Language Models (LLMs) are becoming increasingly powerful for solving many multi-stage problems. One such task is forecasting, a real-world domain that requires a diverse set of capabilities, including information retrieval, tool-use, reasoning, and judgment. For researchers interested in the capabilities of LLMs, forecasting as a domain allows evaluation on all these areas on real-world problems, leveraging the practice of tightly resolving and scoring questions. Forecasting is also extremely difficult, with no clear upper bound on the quality of research and reasoning. As such, it provides a continuous measurement of LLM capabilities: A benchmark based on forecasting will likely never fully be solved.

In this paper, we present Bench to the Future (BTF) a benchmark that assesses LLM capabilities based on their ability to make accurate forecasts. Several frameworks for evaluating AI forecasting have been proposed to date. \cite{halawi2024approaching} provides a rigorous evaluation of AI forecasters using a custom forecasting pipeline on a large, high-quality set of forecasting questions. \cite{schoeneggerWisdomSiliconCrowd2024} analyzed the ability of an ensemble of a "silicon crowd" consisting of twelve LLMs to make accurate prospective forecasts on a set of 31 questions. While they found the LLM ensemble to be indistinguishable from a human crowd, they likely lacked the statistical power to detect meaningful differences in performance. Due to its probabilistic nature, a large number of samples is required to reliably detect differences in forecasting performance \cite{karger2025forecastbench, bosseComparingTwoForecasters2023}. \cite{phanLLMsSuperhuman2024} evaluate LLMs ability to pastcast on a set of 177 Metaculus questions that already resolved. Powered by GPT-4o, their forecasting system has a single evidence gathering prompt asking for a search query. Equipped with evidence returned from that query, a final "forecast" is elicited via a second forecasting prompt. The authors likely relied on Google's date indexing functionality to make sure no evidence after the cutoff date is returned, which is known to be faulty, resulting in issues with data contamination. \cite{hsiehReasoningToolsHuman2024} evaluate a set of agents on their ability to forecast questions from Manifold Markets with a two-week forecasting horizon. Using user-generated questions on Manifold, rather than sourcing them from Metaculus, may have led to the inclusion of questions with trivial topics or unreliable resolutions. Similarly to \cite{phanLLMsSuperhuman2024}, \cite{hsiehReasoningToolsHuman2024} may have suffered from data contamination due to their reliance on the Google Search data cutoff feature.

The analyses mentioned so far were not explicitly meant to be repeated or continuously updated. To this date, two continuously updated benchmarks exist that measure AI forecasting ability: The Metaculus AI Forecasting Benchmark Series \cite{metaculusAIBenchmarking2024} is a quarterly tournament in which bots and professional forecasters compete on hundreds of live, high quality questions. ForecastBench \cite{karger2025forecastbench} is a benchmark featuring an extensive set of forecast questions. All forecasts in these two benchmarks are prospective. This eliminates the risk of data contamination, but also results in long waiting times between forecast and resolution and makes it impossible to run repeated experiments.

BTF, in contrast, is a "pastcasting" benchmark, meaning that resolutions of forecasting questions are known, but LLMs "forecast" on them from the perspective of the past. For every question, we provide a "RetroSearch" environment following from our previous work on Deep Research Bench \cite{futuresearch2025deepresearchbench}. BTF, to this date, comprises a set 299 diverse questions sourced from reputable prediction platforms such as Metaculus \cite{metaculusForecastingPlatform}. It provides a hermetic "pastcasting" environment based on snapshots of the internet with tens of thousands of pre-scraped web pages that allows LLM agents to extensively research questions without concerns of information leakage. This enables  fast evaluations, drastically reducing the time between running the benchmark and obtaining results, and allowing for repeated evaluations of information-retrieval, tool-use, reasoning, and judgment.

In the following we present the benchmark, as well as results from running a set of LLMs and AI forecasting approaches on it. In Section \ref{sec:methods}, we detail our benchmarking methodology, describing the LLMs and AI forecasting architectures we used (Sections \ref{sec:methods-forecasting-approaches} and \ref{sec:methods-models}), our question sourcing and web snapshotting system architecture (\ref{sec:methods-benchmark-setup}), the analyses we performed (\ref{sec:methods-analyses}), and our scoring methodology (\ref{sec:methods-scoring}). In Section \ref{sec:results}, we present the results of our analyses, including our analysis of inter-run variation \ref{sec:methods-inter-run-variance}, performance of LLMs on the main benchmark \ref{sec:results-benchmarking}, and an analysis of the validity of RetroSearch approach based on a comparison between live and retro forecasts \ref{sec:methods-live-vs-retro}. Finally, we discuss our results in Section \ref{sec:discussion}.

\section{Methods}
\label{sec:methods}

\subsection{Benchmark setup}
\label{sec:methods-benchmark-setup}

\begin{figure}[h]
  \centering
  \includegraphics[width=0.8\textwidth]{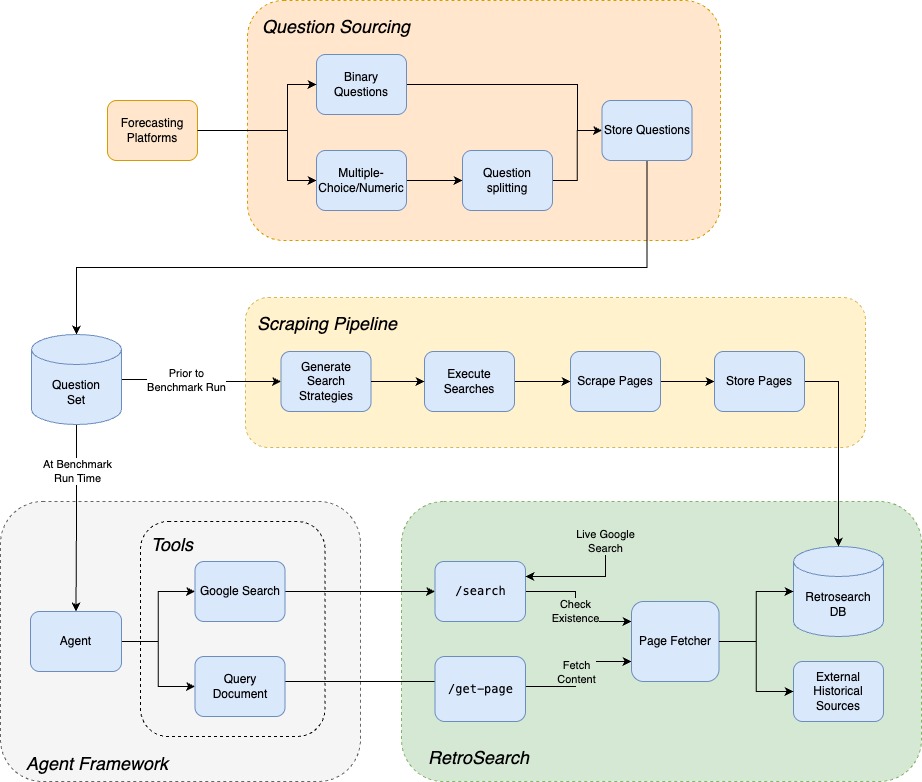}
  \caption{System architecture of Bench to the Future using pastcasting. This covers: question sourcing, snapshot generation per question, and running the benchmark with forecasting agents. (Non-agentic forecasters are omitted for clarity.)}
  \label{fig:retrosearch-arch}
\end{figure}

This section gives an overview of the overall system architecture of Bench to the Future (see Figure \ref{fig:retrosearch-arch}). At the core of this system are question sourcing, and the RetroSearch environment (developed and discussed in detail in \cite{futuresearch2025deepresearchbench}).

\subsubsection{Question Sourcing}
\label{sec:methods-question-sourcing}

BTF sources questions from forecasting platforms, with particular focus on tournament questions on Metaculus. These are particularly suitable due to their quantity and relatively short and predictable resolution time frames. We source binary questions, where a forecaster is asked to predict a single probability of a "yes" outcome to question; multiple-choice questions, where users must spread probability over multiple choices, such as predicting the winner of an election; and numeric questions, where users must predict a distribution over a numeric quantity, such as the number of parameters reported for an upcoming next-generation AI model.

To keep the construct simple for agents and to simplify scoring, we expand all questions into binary questions. We expand multiple-choice questions into $N$ binary questions, where $N$ is number of choices, and expand numeric questions into three binary questions by picking two sensible points within the overall range and then formulating binary questions about resolution below the lower point, above the higher point, and between the two points. This approach does lead to questions that are highly correlated with each other, reducing the overall statistical power of the dataset. We attempt to mitigate this by assigning weights to questions to down-weight questions that are correlated by construction (see details in Section \ref{sec:methods-scoring}, where we discuss our scoring methodology).

BTF comprises a set of 299 binary questions, 87 of them from original binary questions, 135 from multiple-choice questions, and 77 from numeric questions (see Table \ref{tab:question-breakdown}). The effective number of questions (manually down-weighting questions that are correlated by construction) is 178.8. We intend to grow the number of questions over time, and consequently improve the power of our benchmark.

As of this writing, this initial set is sourced from questions that were written between January and April 2025, and resolved between February and April 2025. Future versions of BTF will include more up-to-date questions.

\begin{table}[h]
  \renewcommand{\arraystretch}{1.2}
  \small
  \centering
  \begin{tabular}{lrrr}
    \toprule
     & \textbf{Sourced} & \textbf{Processed} & \textbf{Resolved Positively} \\
    \midrule
    Binary & 87 & 87 & 27 \\
    Multiple-Choice & 24 & 135 & 24 \\
    Numeric & 26 & 77 & 26 \\
    \midrule
    Total & 137 & 299 & 77 \\
    \midrule
    Effective Size & & 178.8 \\
  \end{tabular}
  \vspace{0.5cm}
  \caption{BTF question counts by type and resolution, accounting for processing multiple-choice and numeric questions into binary questions. This includes an effective size, which accounts for questions down-weighted due to inter-question correlations. See Equation \ref{eq:sample-weights} for the weighting scheme. Note that one processed numeric question is missing, which is due to a range being produced that was already trivially resolved at the time of snapshot generation.}
  \label{tab:question-breakdown}
\end{table}

\subsubsection{Snapshot Generation and RetroSearch}
\label{sec:methods-snapshot-generation-retrosearch}

BTF uses a hermetic evaluation environment for forecasting agents called RetroSearch (presented in detail in our previous work \cite{futuresearch2025deepresearchbench}). This approach is based on a large collection of "snapshots" of the web for each question, taken at the time of question sourcing.

We produce snapshots by executing an intelligent web crawl that attempts to exhaustively search over the avenues a forecaster might take when researching a question. These include exploring multiple base rates, establishing the influential actors, and then performing some deep research on these. The end result is a set of around 20,000 pages per question, which we then scrape and store in a database for searching and reading at evaluation time. Table \ref{tab:snapshot-stats} presents some statistics on the number of crawled URLs per question. For split multiple-choice and numeric questions, we typically reuse a single snapshot taken for the original question.

\begin{table}[h]
  \renewcommand{\arraystretch}{1.2}
  \small
  \centering
  \begin{tabular}{lr}
    \toprule
    \textbf{Metric} & \textbf{Value} \\
    \midrule
    Total Questions & 299 \\
    Median Crawled URLs & 21,414 \\
    Crawled URLs $q1$ & 14,390 \\
    Crawled URLs $q3$ & 30,850 \\
    \bottomrule
  \end{tabular}
  \vspace{0.5cm}
  \caption{Snapshot statistics for RetroSearch snapshots used in for BTF benchmark set. $q1$ and $q3$ refer to the first and third quartiles of the number of crawled URLs per question, respectively. Note that some questions share the same snapshot due to being sufficiently correlated.}
  \label{tab:snapshot-stats}
\end{table}

As the snapshots we produce represent a view of the open-web at the date they were produced, they also define a date that we pastcast from. When running pastcasting, we select an available snapshot, constrain RetroSearch to use only pages from that snapshot, and then tell the forecaster that today's date is the date the snapshot was created. The search mechanism is backed by a live Google search, which we filter to only show information we have stored in the RetroSearch database. Using live Google searches ensures a realistic search experience, which we believe outweighs the potential for information leakage based on the fact that the order of certain search results likely changed over time.

\subsection{Models}
\label{sec:methods-models}

The full set of models used are presented in Table \ref{tab:models-and-tools}. For this work, we focused primarily on Anthropic's Sonnet series of models, aiming to establish the utility of this benchmark in tracking the progress of performance of increasingly capable models within the same family. We have also included Gemini 2.5 as a counterpart state-of-the-art model. With these choices, we span a range of capabilities, as well as reasoning and non-reasoning operating modes.

The official training data cutoff for Claude 4 models, released in late May 2025, is March 2025. However, Claude models are instructed to say their training window cutoff is January 2025, leaving some uncertainty about Claude 4's actual training data cutoff. Knowledge about events in March 2025 could in principle make it unsuitable for this iteration of question set, as questions resolved between February and April 2025. Our results in Section \ref{sec:results-calibration} suggest that Claude 4 has very limited, if any, crystallized knowledge of events after January 2025. We also manually checked a variety of real-live events in March 2025 and could not identify any cases where Claude would admit to being aware of them. We therefore included Claude 4 Sonnet, but note that results should be treated with caution.

In general, as new models are released with training window cutoffs too recent for a given pastcasting set, we will run them only on more recently generated questions. BTF is constantly updated for this reason.

\begin{table}[h]
  \renewcommand{\arraystretch}{1.2}
  \small
  \centering
  \begin{tabular}{llll}
    \toprule
    \textbf{Name} & \textbf{Reasoning Mode} & \textbf{Provider} & \textbf{Spec} \\
    \midrule
    Claude Sonnet 4 (Thinking) & Reasoning & Anthropic & claude-sonnet-4-20250514 \\
    Claude Sonnet 4 & Non-Reasoning & Anthropic & claude-sonnet-4-20250514 \\
    Claude Sonnet 3.7 (Thinking) & Reasoning & Anthropic & claude-3-7-sonnet-20250219 \\
    Claude Sonnet 3.7 & Non-Reasoning & Anthropic & claude-3-7-sonnet-20250219 \\
    Claude Sonnet 3.5 v2 & Non-Reasoning & Anthropic & claude-3-5-sonnet-20241022 \\
    Claude Sonnet 3.5 (Original) & Non-Reasoning & Anthropic & claude-3-5-sonnet-20240620 \\
    Gemini 2.5 Pro & Reasoning & Google (Vertex AI) & gemini-2.5-pro-preview-03-25 \\
  \end{tabular}
  \vspace{0.5cm}
  \caption{LLMs used in this work, with reasoning mode and provider.}
  \label{tab:models-and-tools}
\end{table}

\subsection{Forecasting approaches}
\label{sec:methods-forecasting-approaches}

In conjunction with these models, we use a combination of agentic and non-agentic forecasters (see Table \ref{tab:forecaster-architectures}).

\begin{table}[h]
  \renewcommand{\arraystretch}{1.2}
  \small
  \centering
  \begin{tabular}{lp{6cm}p{6cm}}
    \toprule
    \makecell[tl]{\textbf{Name}} & \makecell[tl]{\textbf{Architecture}} & \makecell[tl]{\textbf{Use-cases}}\\
    \midrule
    \makecell[tl]{ReAct Forecaster} & \makecell[tl]{ReAct agent with a forecast-\\optimized task prompt} &
    \makecell[tl]{Main Benchmark\\Evaluation of inter-run variation\\Live vs. Retro validation \vspace{0.1cm}} \\
    %
    %
    \makecell[tl]{Fixed Evidence Forecaster} & \makecell[tl]{A forecast-optimized prompt\\with a fixed set of web evidence} &
    \makecell[tl]{Main Benchmark\\Evaluation of inter-run variation\\Live vs. Retro validation \vspace{0.1cm}} \\
    \makecell[tl]{Variable Evidence Forecaster} & \makecell[tl]{A chain-of-thought information pipeline\\ and a forecast-optimized prompt} &
    \makecell[tl]{Evaluation of inter-run variation\\Live vs. Retro validation \vspace{0.1cm}} \\
    \makecell[tl]{No Evidence Forecaster} & \makecell[tl]{A forecast-optimized prompt\\without any searching or reading pages} &
    \makecell[tl]{Evaluation of inter-run variation\\Live vs. Retro validation \vspace{0.1cm}} \\
  \end{tabular}
  \vspace{0.5cm}
  \caption{Forecaster architectures and where they are used. We use a mixture of agents and chains-of-thought to search and read from the RetroSearch database. All forecasters are given a forecasting-optimized final prompt to elicit the probabilities to be scored.}
  \label{tab:forecaster-architectures}
\end{table}

\subsubsection{ReAct Forecaster}
"ReAct Forecaster" is an agent that follows the ReAct approach defined by Yao et al. \cite{yao2023reactsynergizingreasoningacting}, which in brief is a Thought-Observation-Action loop, with a single forecasting-optimized prompt to elicit the final forecast. See Appendix \ref{sec:appendix-forecasting-prompt} for the full prompt.

Each agent run is prompted with no additional context other than the question and prompt context, and is therefore responsible for gathering its own evidence for each forecast. We enable evidence gathering by providing agents the RetroSearch web-search and page querying tools, following the same approach taken in our Deep Research Bench work \cite{futuresearch2025deepresearchbench}. RetroSearch represents a modified Google Search that only returns results that were previously scraped from the web. The page querying tool allows agents to obtain information from a specific web page. Agents were provided with a "budget" of up to 30 loops of Thought-Observe-Act. In practice, no agent ever came close to exhausting this budget, with the average agent using less than 10 steps before completing.

Traditional ReAct agents, when applied to thinking models, often run into an issue where the "generate a thought" step triggers safeguards preventing actors from trying to access the model's internal reasoning processes. For that reason we re-use the "implicit-thought" modification \cite{futuresearch2025deepresearchbench} for thinking models, where we skip the thought generation step and instead rely on the model's internal reasoning processes. For these agents, the agent loops is only Observe-Act.

\subsubsection{Non-agentic approaches}

Our non-agentic approaches, "Fixed Evidence Forecaster" and "Variable Evidence Forecaster" use the same forecasting-optimized prompt with a set of approximately 30 pieces of pre-sourced evidence from the RetroSearch database included. "Variable Evidence Forecaster" first runs a chain-of-thought evidence gathering pipeline, expanding the question into around 20 subqueries, searches the RetroSearch database using these queries, reads the resulting webpages, and then extracts a set of around 30 salient facts. Facts are extracted from individual pages using Gemini 2.5 Flash, while the search query generation, and the selection of the most relevant resulting facts is carried out by the model under test. This non-agentic approach is inspired by the observation of tournament-winning performance of simple one-prompt forecasters paired with good evidence gathering in the Metaculus AI Forecasting Benchmark Series \cite{metaculusAIBenchmarking2024}.

For efficiency, we use the Fixed Evidence Forecaster for the main benchmark runs. In contrast to the Variable Evidence Forecaster, it does not run the evidence gathering pipeline every time it makes a forecast. Instead, we did a single evidence gathering run for all questions, using Claude Sonnet 3.7 (plus Gemini 2.5 Flash for fact extraction) and re-used this evidence for all Fixed Evidence Forecaster runs.

As a baseline, we also include a "No Evidence Forecaster". This is the same forecast-tuned prompt but without any data from the RetroSearch database. Using this approach, we can understand the effect of producing evidence, and, crucially, understand any cases where the model "remembers" the answer to any questions. This was especially important when assessing the suitability of including Claude Sonnet 4, as given its March 2025 training window cutoff it may be able to trivially forecast by having the outcomes accessible from its weights. In general, however, this forecast approach is very poor as often the LLMs are not aware of the entities or situation referenced in the question.

\subsection{Analyses}
\label{sec:methods-analyses}

For our main benchmark, we executed 5 runs of the React Forecaster, the Fixed Evidence Forecaster, and the No Evidence Forecaster (with the number 5 chosen experimentally, see Section \ref{sec:methods-inter-run-variance}) on each of the 299 questions in the BTF set. We use the mean of those 5 runs as the final forecast for scoring.

\subsubsection{Inter-run variation}
\label{sec:methods-inter-run-variance}

Forecasting is probabilistic, and LLMs produce different results when prompted with the same forecasting task multiple times. Variation between forecasts needs to be sufficiently low for the results of the benchmark evaluations to be stable. To measure inter-run variation in the benchmark, we ran our forecasting approaches repeatedly on a subset of 18 questions. The 18 questions were selected at random, with a check to ensure none were obviously correlated with another.

For each question, we ran 20 forecasts and visually inspected plots of the variation in the forecasts. The number 20 was chosen to balance the need for a sufficiently large sample with computational cost. In addition, we also ran a bootstrap analysis with $n=1000$ repetitions where, for a given question, we computed the mean of $n$ randomly sampled forecasts (from the 20 available per question). We ultimately chose to use the mean of 5 runs for our main runs based on visual inspection of results, balancing the desire to achieve low variation in final forecast with considerations related to time and compute cost. Results are presented in Section \ref{sec:results-inter-run-variation}, including the variation of the bootstrapped mean across 5 samples.

\subsubsection{Fidelity of Retro forecasts to Live forecasts}
\label{sec:methods-live-vs-retro}

To assess the validity of our pastcasting approach, we analyzed the distribution of live and retro forecasts via a distinct, curated set of 20 questions. These are 20 questions that were open at the time of evaluation, and are separate from our BTF benchmark set. We used the ReAct Forecaster, Variable Evidence Forecaster and No Evidence Forecaster here as we did for the inter-run variation analysis, and ran every approach 20 times for each question in both live and retro mode. The No Evidence Forecaster serves as a baseline, as we expect these runs to be identical for live and retro mode. This allows to gauge the level of noise inherent in this analysis, therefore aiding us in making meaningful comparisons for evidence-utilizing forecasters. The results of this analysis are discussed in Section \ref{sec:results-live-vs-retro}.

\subsection{Scoring}
\label{sec:methods-scoring}

Our primary metric for assessing forecasting performance is the Brier score \cite{brierVerificationForecastsExpressed1950}, which measures the squared difference between the forecast and the actual outcome:

\begin{equation}
  \text{Brier score}_i = \frac{1}{N} \sum_{i=1}^{N} (y_i - f_i)^2
\end{equation}

where $y_i$ is the actual outcome and $f_i$ is the forecast.

The questions in our benchmark are correlated, both by construction and by the fact that they cover a limited range of topics. This reduces the effective sample size of BTF. Following the approach chosen for the Metaculus AI Benchmarking Tournament \cite{metaculusAIBenchmarking2024}, we account for correlations between questions that are related by construction, i.e. questions that were split into binary questions from multiple choice or continuous questions. For every group of $N_i$ related questions, we assign the following weights:

\begin{equation}
\label{eq:sample-weights}
  \text{weight}_i = \frac{\log_2(N_i+1)}{N_i+1}.
\end{equation}

We then compute the weighted mean and standard deviation of the Brier scores for each forecasting approach. We do not attempt estimating correlations between the original binary questions sourced from Metaculus and therefore do not account for it, assigning everyone of those questions a weight of 1.

In the process of scoring, we noticed 11 anomalous individual forecasts that produced values >1. The primary failure mode is that, despite our best prompting efforts, very occasionally a model will respond with a percentage $\exists [0, 100]$ rather than a probability $\exists [0, 1]$. The secondary failure mode is that some split numeric questions confuse the forecaster, causing them to respond with the numeric value rather than a probability within the stated range. Given how low ($\simeq 0.003\%$) the occurrence of these cases is, we chose to filter these points out rather than attempt to correct them.

\section{Results}
\label{sec:results}

\subsection{Inter-run variation}
\label{sec:results-inter-run-variation}

\begin{figure}[h]
  \centering
  \includegraphics[width=0.99\textwidth]{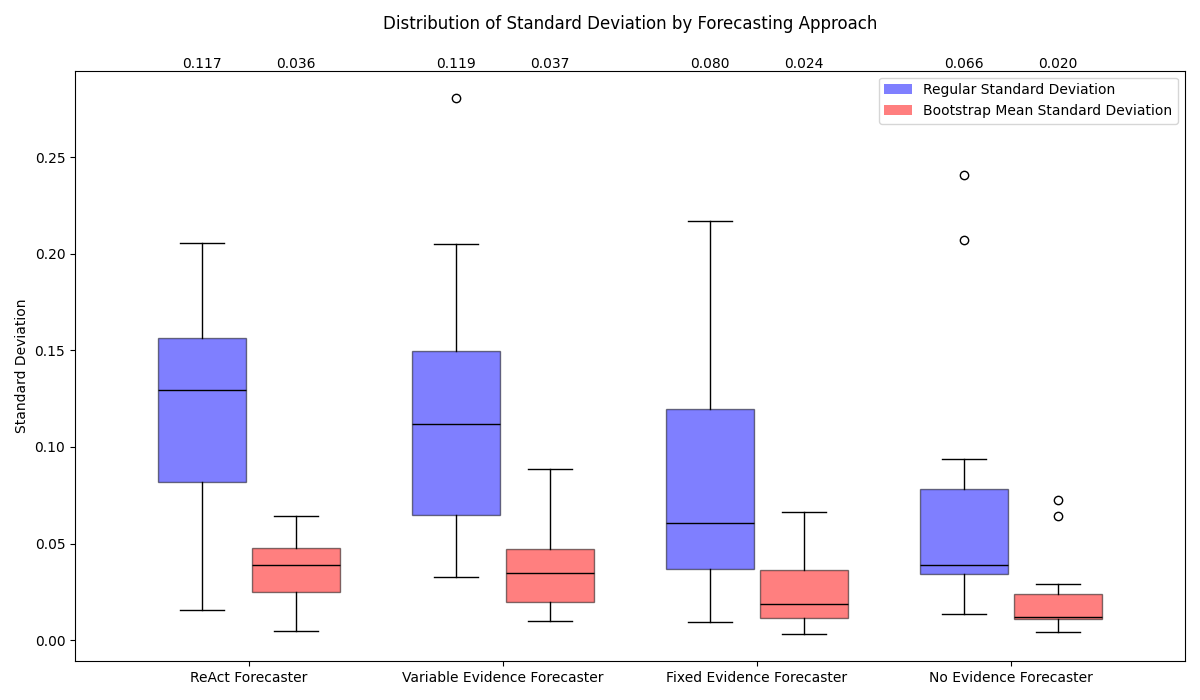}
  \caption{Inter-run variation by standard deviation of 20 forecasts from each approach on each of 18 questions. All approaches were powered by Gemini 2.5 Pro. Red boxes show the standard deviation obtained by bootstrapping 5 forecasts from the available 20 per question (with 1000 iterations).}
  \label{fig:inter-run-variation-boxplot}
\end{figure}

\begin{figure}[h]
  \centering
  \includegraphics[width=0.99\textwidth]{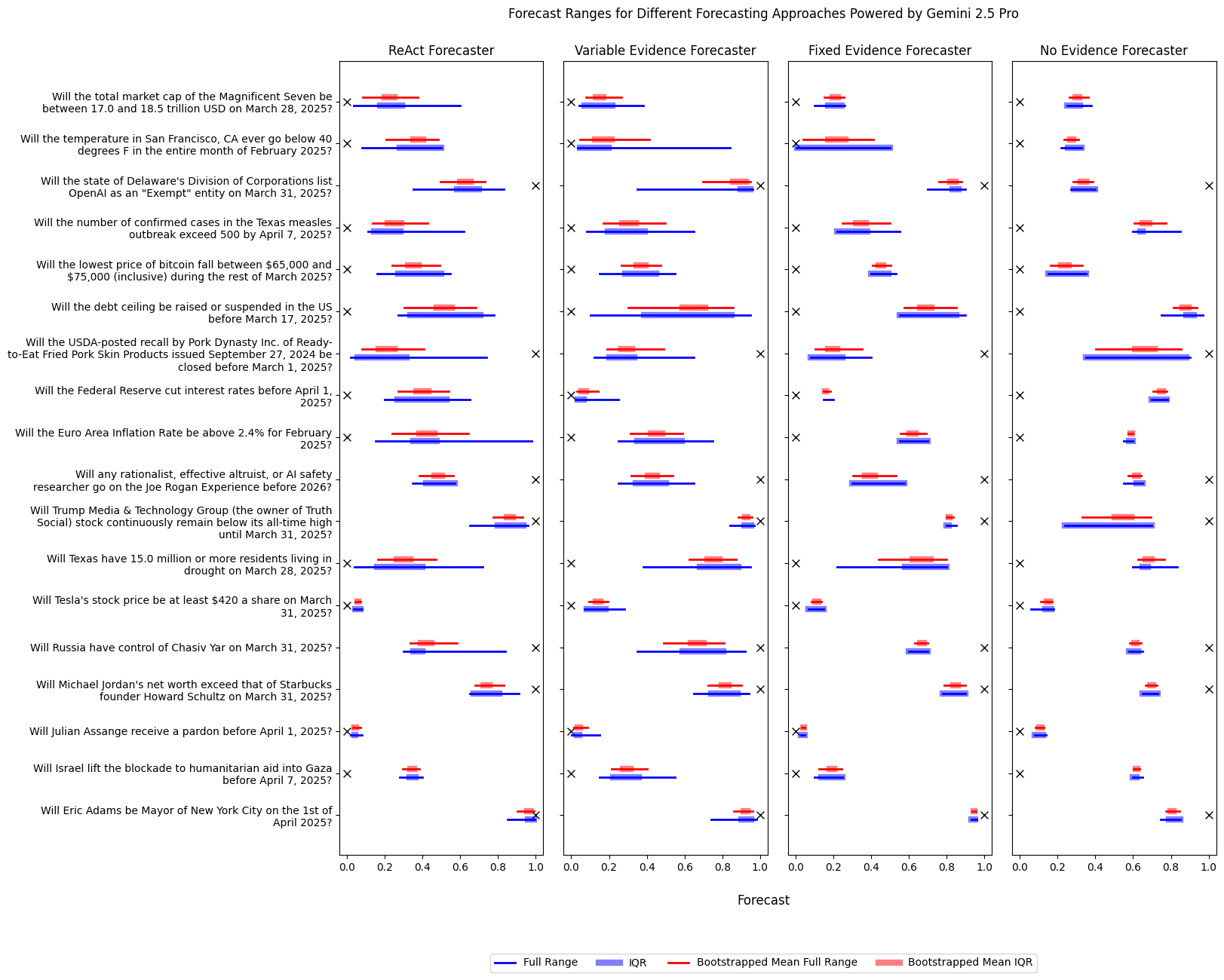}
  \caption{Inter-run variation shown as full ranges (line) and interquartile ranges (box) of forecasts for different forecasting approaches on a set of 18 questions. All forecasts were powered by Gemini 2.5 Pro. Ranges for bootstrapped means of 5 forecasts are shown in red, ranges for regular forecasts in blue. Resolutions (yes = 1, no = 0) are marked with a black x.}
  \label{fig:inter-run-variation-range}
\end{figure}

The range of standard deviations for each forecaster is shown in Figure \ref{fig:inter-run-variation-boxplot}. We see that 20 repeat forecasts on a set of 18 question showed substantial inter-run variation across all forecasters. For a significant number of questions from our test set, individual forecasts exhibited a very wide range of possible values (see Figure \ref{fig:inter-run-variation-range}), sometimes giving probabilities $<20\%$ on one run and $>80\%$ on another. This suggests that single forecasts from an LLM should not be considered reliable. The forecast variation can be reduced by averaging multiple runs: transitioning from single runs to the average of 5 runs consistently decreased the standard deviation by approximately two-thirds across all approaches.

When comparing the Fixed Evidence Forecaster and the Variable Evidence Forecaster, we can see that the standard deviation of the variable evidence approach was approximately 1.7 times higher than that of the fixed evidence approach. This suggests that, in our sample, and for this particular approach with a final forecasting prompt that relies on pre-sourced evidence, approximately $60\%$ of the variation stemmed from the final judgment prompt, while only $40\%$ originated from the information gathering process. Interestingly, the standard deviation observed in the Fixed Evidence Forecaster exceeded that of the No Evidence Forecaster. Adding information (or, perhaps, just lengthening the prompt) evidently increased inter-run variation.

\subsection{Benchmark results}
\label{sec:results-benchmarking}

\begin{figure}[h]
  \centering
  \includegraphics[width=0.99\textwidth]{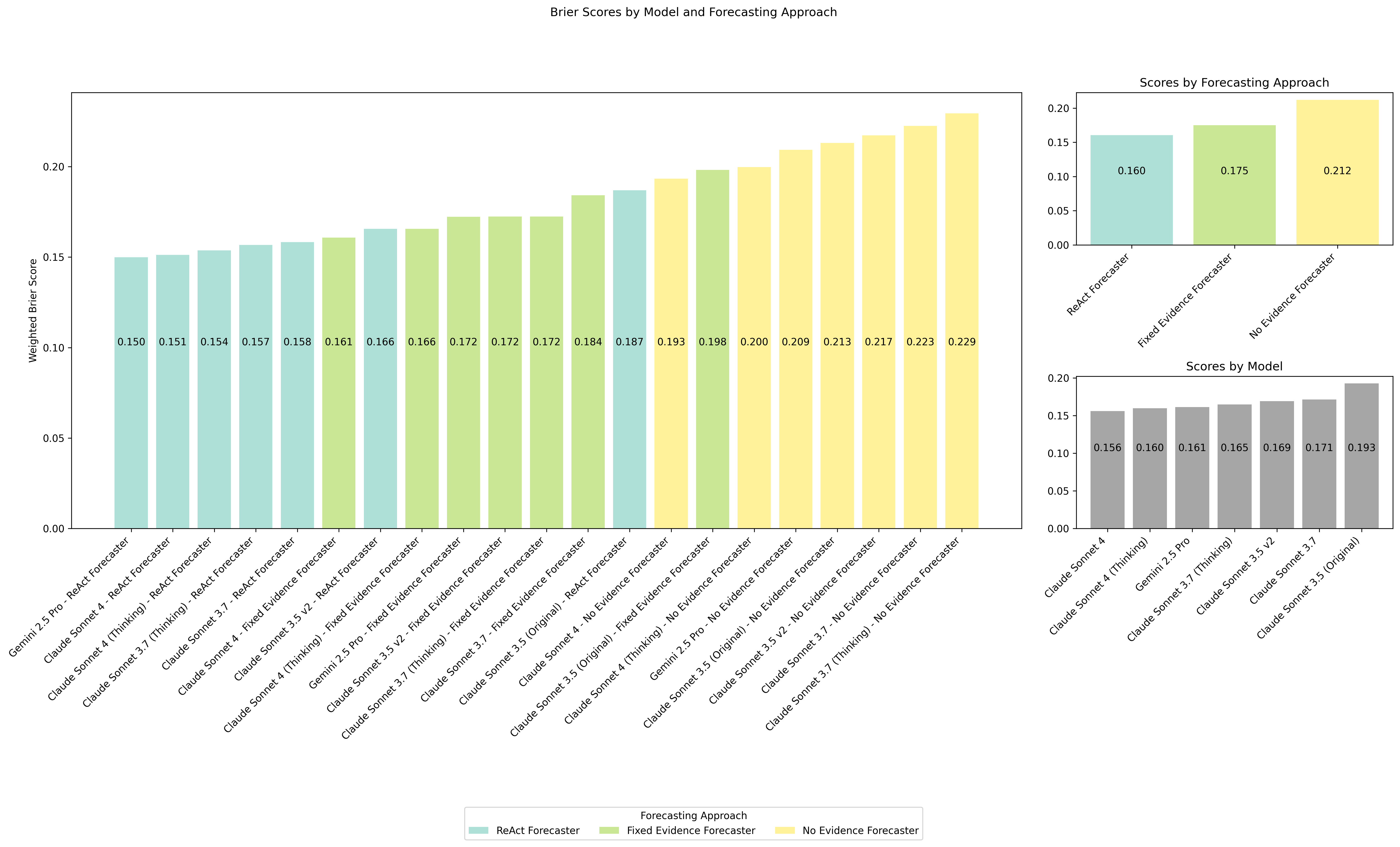}
  \caption{Brier scores for forecasting approaches and models. Bars represent weighted means across all forecasting questions. Scores for agents/models are averaged across models/agents. To compute model score, we only took the ReAct Forecaster and the Fixed Evidence Forecaster into account, ignoring the No Evidence baseline.}
  \label{fig:brier-scores-benchmark}
\end{figure}

We present the results of our full benchmark evaluation in Figure \ref{fig:brier-scores-benchmark}.

\subsubsection{Accuracy scores across LLMs}
\label{sec:results-llms}

The data shows a steady accuracy improvement between older models and newer models, in line with improvements to other benchmarks of these same models. We find that Claude Sonnet 4 and Gemini 2.5 Pro typically outperform Claude Sonnet 3.7, which is itself significantly better than Claude Sonnet 3.5 v2 and Claude Sonnet 3.5 (Original). We also find that the agentic ReAct Forecaster approach entirely outperforms its counterpart Fixed Evidence Forecaster per model, with the best performing forecaster being the Gemini 2.5 Pro ReAct Forecaster.

Table \ref{tab:model-comparisons} shows results from paired t-tests based on pairwise comparisons between models. Despite not correcting for for multiple testing, most comparisons are not significant at a conventional 5\% significance level, underlining the large number of observations required to achieve statistical power. However, comparisons show results that are clearly in line with prior belief, suggesting that we're observing a real effect, rather than noise.

Newer models generally have more up-to-date information. For a question about events in Feb 2025, for example, it's plausible that one model with crystallized knowledge about Dec 2024 would outperform one without such knowledge, even though it is months out of date to the question at hand. Since this is a confounding variable we don't control for, we cannot confidently state that newer LLM outperforming older LLMs is entirely due to planning, tool-use, reasoning, and judgment.

We don't find significant accuracy improvements of "thinking" models vs "non-thinking" models. For ReAct, this is consistent with Deep Research Bench \cite{futuresearch2025deepresearchbench}, where the "thinking" done at each step is satisfactorily replaced by an additional step for non-thinking models to output their thoughts before choosing their actions. However, for the Fixed Evidence forecaster, as well as in the final forecasting prompts, there could plausibly be a high return to additional "thinking", but we do not observe that here.

While we can compare models executed on this benchmark to each other, absolute Briers scores are hard to interpret. No human baseline is available at present, and different sets of forecasting questions have different difficulty levels.

\begin{table}[h]
  \centering
  \begin{tabular}{llrrr}
    \toprule
    Model 1 & Model 2 & Brier Score 1 & Brier Score 2 & p-value \\
    \midrule
    Claude Sonnet 3.5 (Original) & Claude Sonnet 4 & 0.1925 & 0.1560 & 0.0024 \\
    Claude Sonnet 3.5 (Original) & Claude Sonnet 3.7 (Thinking) & 0.1925 & 0.1645 & 0.0024 \\
    Claude Sonnet 3.5 (Original) & Claude Sonnet 4 (Thinking) & 0.1925 & 0.1597 & 0.0055 \\
    Claude Sonnet 3.5 (Original) & Claude Sonnet 3.5 v2 & 0.1925 & 0.1690 & 0.0198 \\
    Claude Sonnet 3.5 (Original) & Gemini 2.5 Pro & 0.1925 & 0.1611 & 0.0230 \\
    Claude Sonnet 3.5 (Original) & Claude Sonnet 3.7 & 0.1925 & 0.1712 & 0.0266 \\
    Claude Sonnet 3.7 & Claude Sonnet 4 & 0.1712 & 0.1560 & 0.1393 \\
    Claude Sonnet 3.7 & Claude Sonnet 3.7 (Thinking) & 0.1712 & 0.1645 & 0.1696 \\
    Claude Sonnet 3.7 & Claude Sonnet 4 (Thinking) & 0.1712 & 0.1597 & 0.2017 \\
    Claude Sonnet 3.5 v2 & Claude Sonnet 4 & 0.1690 & 0.1560 & 0.2275 \\
    Claude Sonnet 3.7 & Gemini 2.5 Pro & 0.1712 & 0.1611 & 0.3536 \\
    Claude Sonnet 3.5 v2 & Claude Sonnet 4 (Thinking) & 0.1690 & 0.1597 & 0.3585 \\
    Claude Sonnet 3.7 (Thinking) & Claude Sonnet 4 & 0.1645 & 0.1560 & 0.3676 \\
    Claude Sonnet 3.5 v2 & Gemini 2.5 Pro & 0.1690 & 0.1611 & 0.4871 \\
    Claude Sonnet 4 & Claude Sonnet 4 (Thinking) & 0.1560 & 0.1597 & 0.5292 \\
    Claude Sonnet 3.7 (Thinking) & Claude Sonnet 4 (Thinking) & 0.1645 & 0.1597 & 0.5343 \\
    Claude Sonnet 3.5 v2 & Claude Sonnet 3.7 (Thinking) & 0.1690 & 0.1645 & 0.6137 \\
    Claude Sonnet 4 & Gemini 2.5 Pro & 0.1560 & 0.1611 & 0.6224 \\
    Claude Sonnet 3.7 (Thinking) & Gemini 2.5 Pro & 0.1645 & 0.1611 & 0.7337 \\
    Claude Sonnet 3.5 v2 & Claude Sonnet 3.7 & 0.1690 & 0.1712 & 0.8184 \\
    Claude Sonnet 4 (Thinking) & Gemini 2.5 Pro & 0.1597 & 0.1611 & 0.8811 \\
    \bottomrule
  \end{tabular}
  \caption{Pairwise comparisons between models using paired t-tests (taking question weights into account). For the pairwise comparisons, we averaged across repeat runs for a single question/approach, then averaged scores across approaches, resulting in one score per model and questions. We only took scores for the ReAct Forecaster and the Fixed Evidence Forecaster into account, ignoring the No Evidence baseline. P-values are not adjusted for multiple comparisons.}
  \label{tab:model-comparisons}
\end{table}

\subsubsection{Accuracy scores across forecasting approaches}
\label{sec:results-approaches}

The No Evidence Forecaster scores show markedly worse forecasts than approaches that were able to gather evidence before making a forecast. Yet, all No Evidence Forecasters performed better than a naive baseline of predicting 50\% on every question (resulting in a Brier score of 0.25), with the best models, Claude 4 with and without thinking, being noticeably better. This suggests that there may be returns to better judgment, even in the absence of any specific information about question at hand. Note, however, that results are confounded by the amount of crystallized information LLMs have access to. This is true in particular for Claude 4 and its more recent cutoff window, as mentioned in Section \ref{sec:methods-models}.

In terms of approaches with access to research on the specific question (or the ability to conduct that research autonomously), the ReAct Forecaster approach performed better than the Fixed Evidence Forecaster. We hypothesize that ReAct forecasters accessed fewer websites and likely collected fewer total evidence items, but made up for the reduced breadth with their ability to strategically select and collect evidence. The inability of our hard-coded pipeline to return sufficiently relevant information may be an artifact of our specific implementation. But it otherwise suggests a return to depth-based approaches compared to breadth-based approaches: Finding many useful facts up front may not be as useful as finding a small number of facts, evaluating them, and searching more info in an iterative fashion.


\subsubsection{Calibration}
\label{sec:results-calibration}

\begin{figure}[h]
  \centering
  \includegraphics[width=0.99\textwidth]{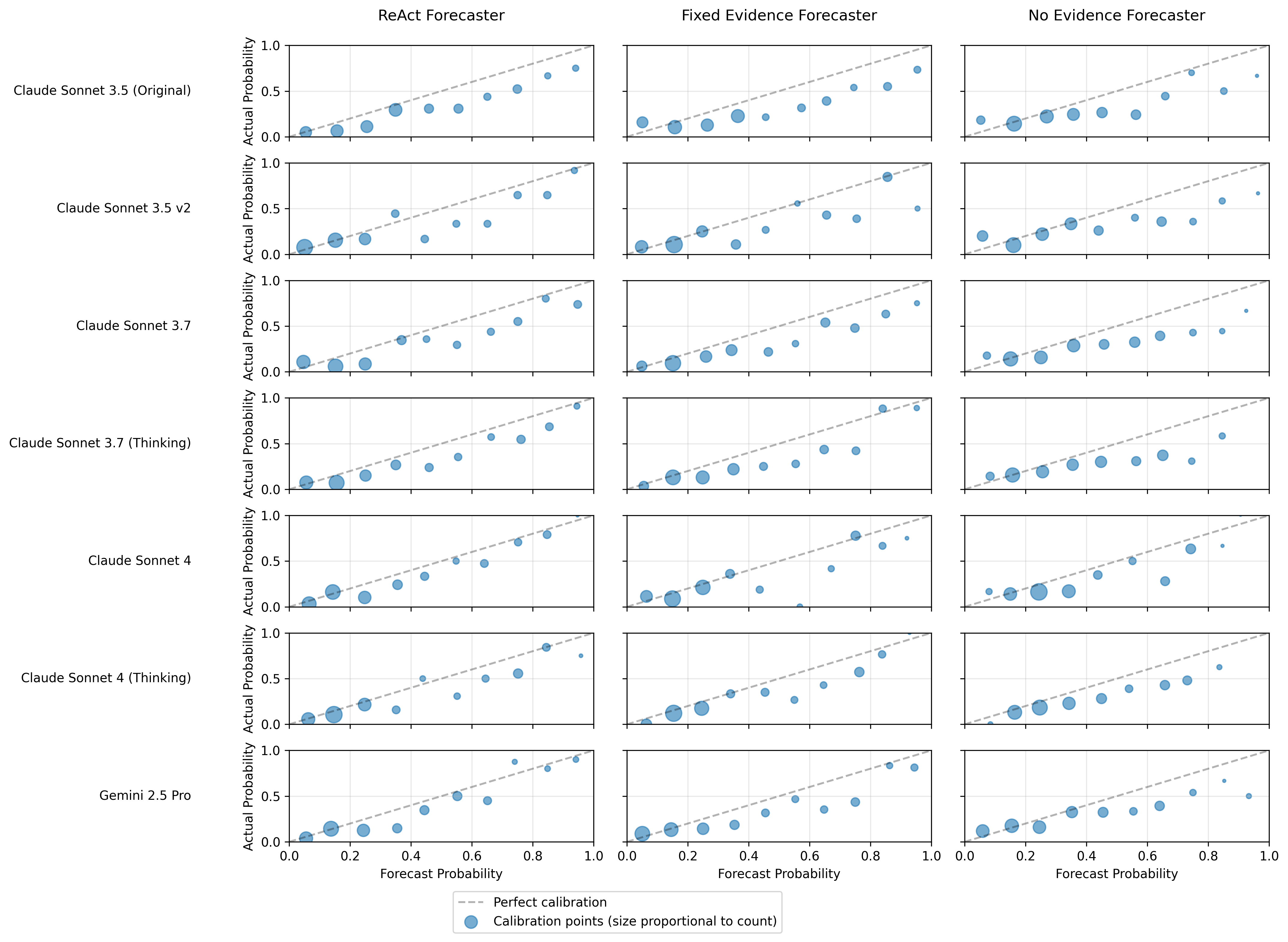}
  \caption{Calibration plots for each combination of model and agent.}
  \label{fig:calibration-benchmark}
\end{figure}

In Figure \ref{fig:calibration-benchmark}, we present forecast calibration plots each combination of model and agent. These relate the probability assigned by the forecaster to the frequency of the outcome occurring across the full range of probabilities. A perfectly-calibrated ideal forecaster would have outcomes predicted with $x\%$ occur $x\%$ of the time for all $x$, so observing deviation from this trend in a real forecaster provides a useful evaluation of how over- or underconfident they are. We find that AI forecasters are not perfectly calibrated, with each forecaster exhibiting some systematic biases in their predictions. Most notably, all models and approaches exhibited a consistent tendency to predict "Yes" outcomes more frequently than the actual resolutions would warrant, as observed in the calibration being below the perfect calibration line for forecasts $>0.5$.

The majority of questions in our dataset resolved as "No" (see Table \ref{tab:question-breakdown}), consistent with a known pattern on forecasting platforms where the majority of questions resolve negatively, and exacerbated by the fact that splitting multiple-choice and numeric questions into binary outcomes means that many more questions must resolve "No" than "Yes". For a well calibrated forecaster, this should not make a difference. For a random number generator, however, a systemic bias towards "No" resolutions in the data would show up as a "Yes" bias, which is what we see here across all models and approaches. Further analysis is needed to discern whether the observed bias is more an artifact of our data or a persistent feature of LLM forecasters. A visualization of forecast resolution can be found in Figure \ref{fig:resolution-benchmark} in the Appendix.

\subsection{Comparing Live and Retro forecasts}
\label{sec:results-live-vs-retro}

\begin{figure}[h]
  \centering
  \includegraphics[width=0.99\textwidth]{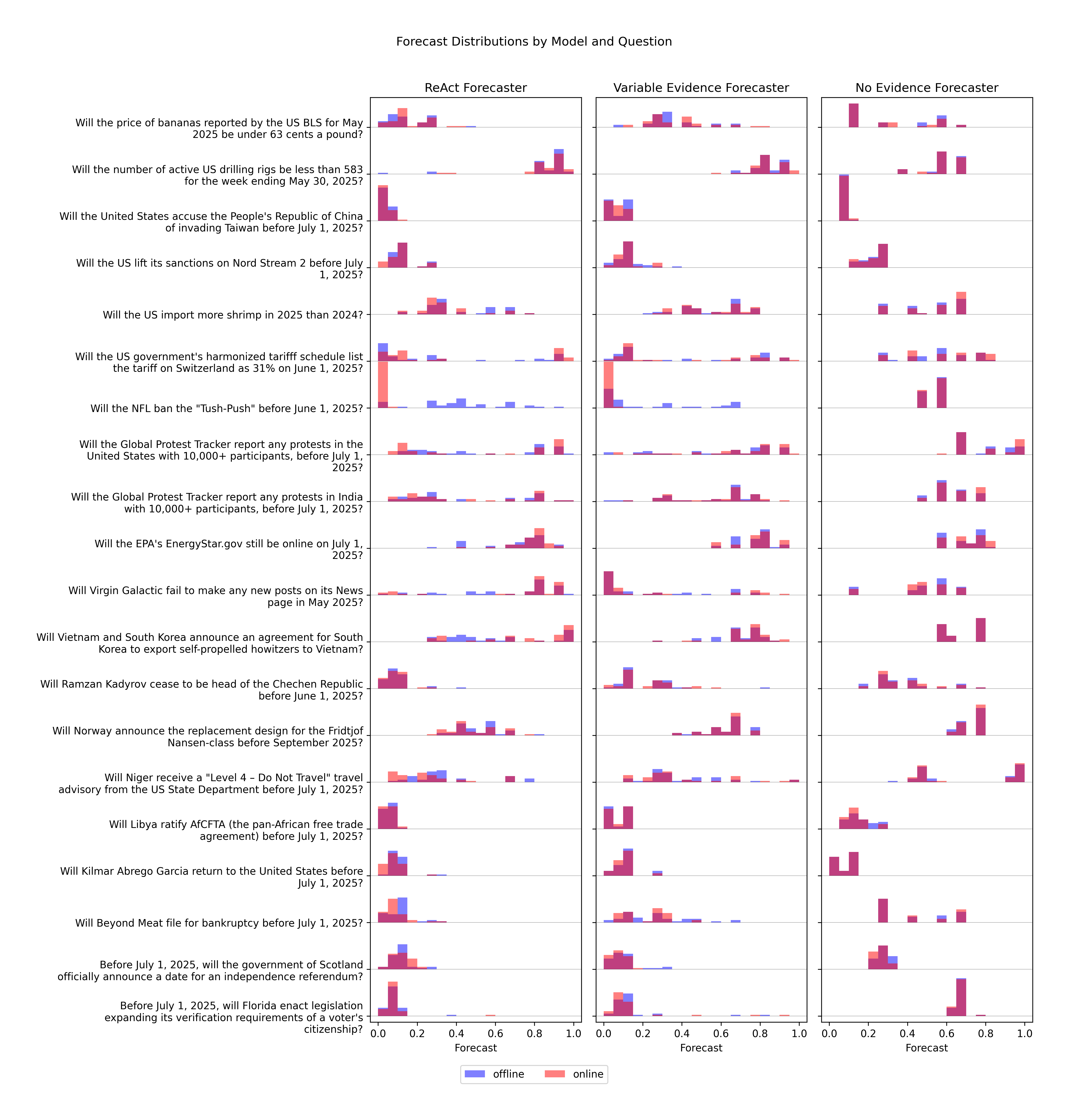}
  \caption{Forecast distributions between approaches with the Live Internet (red) vs. RetroSearch database (blue) by model and question. The x-axis shows the forecast value, and the y-axis shows the frequency of forecasts.}
  \label{fig:live-vs-retro-forecasts}
\end{figure}

The distribution of forecasts using the live internet, as well as those using the RetroSearch database on a separate set of 20 questions is shown in Figure \ref{fig:live-vs-retro-forecasts}. Histograms are based on 20 samples taken per question using ReAct Forecaster, Variable Evidence Forecaster and, as a control, No Evidence Forecaster. Visually, we observe good correspondence between the live and retro distributions of each forecaster in all but one cases. Crucially, differences between these distributions in the ReAct Forecaster and Variable Evidence Forecaster typically do not look much greater than for a No Evidence Forecaster. The No Evidence Forecaster does not access any evidence and hence is not influenced by being in live vs retro mode, therefore representing a reference for expected agreement in an ideal scenario.

In this set, there was only one question with dramatically different forecasts when done in the RetroSearch environment vs using the live internet, the question: "Will the NFL ban the Tush Push before June1, 2025?". On May 21, NFL owners met to discuss the matter and a vote to ban it failed. Scraping for the question was done on May 20, 2025, while the live forecasts were made between May 21 and May 24. The live forecasts ostensibly picked up on the failed vote, while retro forecasts did not, resulting in much higher forecasts. This highlights a failure mode of the approach when trying to compare live against retro forecasts. However, note that this does not automatically invalidate the retro forecasts themselves: differences in performance between models is still meaningful, as long as all forecasting approaches have access to comparable information.

We also observe that RetroSearch forecasts seem to have higher variance on some questions than live-web forecasts. This may be due to similar issues such as the live web having a web page with decisive information about a question that is not in the RetroSearch database. Note that decisive information doesn't mean the best forecast is 0 or 1. For example, the current stock price might be critical information to have on a question of future stock prices, even if the best forecast would remain highly uncertain.

\section{Discussion}
\label{sec:discussion}

In this paper, we present Bench to the Future (BTF), a novel pastcasting benchmark for evaluating the forecasting capabilities of Large Language Models (LLMs). The benchmark enables a reliable, repeatable and rapid way to evaluate LLM forecasting capabilities via its hermetic RetroSearch approach. BTF currently comprises 299 high-quality questions sourced from prediction platforms, with each question accompanied by an extensive offline corpus of relevant web pages. We evaluated three forecasting approaches (ReAct agent, non-agentic forecaster, no-evidence LLM baseline) across several state-of-the-art LLMs, including Claude Sonnet 3.5 (Original and v2), Claude Sonnet 3.7, Claude Sonnet 4, and Gemini 2.5 Pro, in both reasoning and non-reasoning modes.

To validate our pastcasting approach, we conducted a comparative analysis between live and retro forecasts on a separate set of 20 questions. The results indicated that our pastcasting methodology produces forecasts that are broadly consistent with live forecasting, though further validation with a larger sample size is warranted. Inter-run forecast variation could be substantially reduced by averaging across multiple runs. Our comparison between a Fixed Evidence Forecaster and a Variable Evidence Forecaster suggest that a substantial fraction of the overall inter-run forecast variation using these approaches can be attributed to the final forecasting prompt, as opposed to the the evidence gathering step. The main benchmark results suggest a steady increase of performance with every new model release. ReAct Forecasters generally outperformed non-agentic forecasters with a chain-of-thought evidence gathering pipeline, which in turn outperformed No Evidence Forecasters.

We acknowledge a number of limitations with this initial version of BTF. The first one is the sample size. While results correspond well to intuitions about relative performance between models, more questions are needed to discern differences between forecasters with conventional significance levels. Furthermore, the existing question set will soon be out of date, with new models being published with more recent training cutoff dates. Scraping of questions beyond this initial 299 question set are in progress, and we will publish updated versions that allow for testing of LLMs with more recent training window cutoffs, as well as having a higher statistical sample size.

While preliminary analyses suggest good correspondence between forecasts in live and retro mode, the approach comes with a number of fundamental limitations. RetroSearch may bias forecasts in ways that are hard to detect and require more analysis to discover. Not that biases do only invalidate the approach insofar they affect different models differently: as long as the relative rankings between models stay the same between live and retro mode, we can still make valid inferences. More research is necessary to validate this, in particular comparing scores and rankings obtained in live and retro mode. As LLMs become more capable, our benchmark is prone to underestimating performance of the best forecasters: It is possible the future agents will explore avenues our current scraper does not adequately anticipate, and this might hinder our ability to capture performance improvement. One concern with the current implementation of RetroSearch is the potential of information leaking due to our reliance on live Google searches to fetch websites from our RetroSearch database (returning, of course, only previously scraped web pages). We aim to rectify in future incarnations.

Our approach enables us to perform agent-to-agent comparison, and track the progression of capability with new models and agent approaches. One crucial piece that is missing at present is positioning their performance against a human baseline. We cannot interpret absolute Brier scores on their own terms, as there is no way to disentangle forecasting ability from question difficulty. Our initial benchmark has been limited by not being able to adequately scrape content at times comparable to the initial forecast window of our source questions, which means often when a human baseline could be obtained, we have not yet been able to. This is a technical challenge we expect to solve in future iterations of the benchmark.

The approach we have produced is a powerful and novel approach to grading AI forecasters. Not only do we have opportunities to expand the significance and discriminative capabilities in grading raw forecasting ability, we can also explore many other facets of forecasting that we have not considered here. For example, we can explore consistency analysis, where explore a complete set of possibilities for groups of binary questions and gauge how close the sum of forecasts are to 1. Similarly, we have yet to explore the yes/no bias of agents, which we can do by intelligently flipping the direction of a set of questions and comparing the original forecast to the inverse. We are also interested in exploring the details of the decision making processes of agents using the trace evaluation approach applied for DRB \cite{futuresearch2025deepresearchbench}.


\section*{Acknowledgments}
This work was supported by Open Philanthropy.

\printbibliography

@article{brierVerificationForecastsExpressed1950,
  title = {Verification of {{Forecasts Expressed}} in {{Terms}} of {{Probability}}},
  author = {Brier, Glenn W.},
  year = {1950},
  month = jan,
  journal = {Monthly Weather Review},
  volume = {78},
  number = {1},
  pages = {1--3},
  publisher = {American Meteorological Society},
  issn = {1520-0493, 0027-0644},
  doi = {10.1175/1520-0493(1950)078<0001:VOFEIT>2.0.CO;2},
  urldate = {2022-01-21},
  chapter = {Monthly Weather Review},
  langid = {english}
}

\clearpage

\appendix
\section{Appendix}
\label{sec:appendix}

\subsection{Forecasting prompt}
\label{sec:appendix-forecasting-prompt}

The following prompt was used for all forecasting runs (we added the question to the prompt, but left the evidence as a placeholder).

\begin{tcolorbox}[
    breakable,
    width=\textwidth,
    left=10pt,
    right=10pt,
    top=6pt,
    bottom=6pt,
    boxrule=0.5pt,
    colback=gray!10,
    colframe=gray!50,
    title=Forecasting Prompt
]
\begin{lstlisting}[
    breaklines=true,
    breakatwhitespace=true,
    basicstyle=\small\ttfamily,
    frame=single,
    framesep=2mm,
    xleftmargin=2mm,
    xrightmargin=2mm,
    backgroundcolor=\color{gray!10},
    showstringspaces=false,
    columns=flexible,
    literate={https://}{https://\allowbreak}1
            {http://}{http://\allowbreak}1
            {www.}{www.\allowbreak}1
            {.com}{.com\allowbreak}1
            {.org}{.org\allowbreak}1
            {.net}{.net\allowbreak}1
            {.edu}{.edu\allowbreak}1
            {.gov}{.gov\allowbreak}1
            {/}{/\allowbreak}1
            {?}{?\allowbreak}1
            {&}{&\allowbreak}1
            {=}{=\allowbreak}1
]
You are a professional forecaster interviewing for a job.

Your interview question is:

You are tasked with giving a probabilistic forecast for the following question:
<question>
Will the Joe Rogan Experience be ranked 1st on the Spotify Podcast Charts on March 31, 2025?
</question>

Here is some background on the question that you may find helpful:
<background>
[American Best-Selling Author Beats Joe Rogan's $250M Spotify Podcast for a 4th Consecutive
Top Spot](https://www.essentiallysports.com/ufc-mma-news-american-best-selling-author-beats-
joe-rogans-two-fifty-million-dollar-spotify-podcast-for-a-fourth-consecutive-top-spot/)

Ensure settings are on United States and Top Podcasts
</background>

The following criteria specify precisely how the question's resolution will be determined:
<resolution-criteria>
This question resolves as YES if The Joe Rogan Experience is ranked exactly 1st on the Spotify Podcast Charts at [this link](https://podcastcharts.byspotify.com/) when checked by Metaculus on or after March 31, 2025, and NO otherwise.
</resolution-criteria>

Please use the information and research gathered by your trusted assistant below:
<research>
{evidence_output}
</research>

Today is {self.present_date.strftime("%Y-%m-%d")}.

Here are top tips from good forecasters:

If an event was anticipated to happen in a certain timeframe, but 80% of that time has passed and there is no recent news or updates about it happening soon, then you should be skeptical that it will happen on the originally stated timeframe. It probably means that it will be delayed or plans have changed. Some examples:
If Elon Musk says definitely that robotaxis will be available in 1 year, but after 11 months there have not been any specific public updates confirming that robotaxis will be available then it is highly unlikely to happen in the remaining month.
If Donald Trump says that he will definitely have a deal signed in 90 days, but 70 days have passed without updates on the progress, then it is highly unlikely there will be a deal in the remaining 20 days.

Think about base rates for similar events in the past. Sometimes a base rate is the best you can do, if you can't find much information about the question. Example questions where the base rate is a particularly good starting point include:
Will the temperature in Miami exceed 100 degrees next month?
Will there be an earthquake of magnitude 5 or more in San Francisco in 2026?

Sometimes finding a good base rate is difficult, especially when the events are relatively unique. In these cases, you need to put more weight on the "inside view" which means weighing considerations that appear specific to the situation that lead to a very different forecast than base rates would dictate. You will need to use your own judgment.

Put extra weight on the status quo outcome since the world changes slowly most of the time. This is especially true when coordination or agreement between people or organizations is required. For example, signing multi-national treaties and passing legislation often take longer than one might imagine from reading the news.

Think about if there are seasonal effects. For example, the sales of homes or travel are likely to have seasonal patterns.

Think about what the current trend is and if it makes sense to extrapolate, or not. Some things like stock prices are effectively random walks, so recent trends likely don't matter. Other trends have momentum, like the number of COVID cases from day to day.

Think about the scope of the question.

Think about the incentives and power of any influential people involved in the situation. For example, Putin has the power to single-handedly dictate Russian military or diplomatic response.

Sometimes there are multiple data sources for the same number that have very different values. For example, Trading Economics reports 1.84B UAH for Ukrainian debt while the IMF reports a value of 7B UAH. These are both reputable sources, but they are using different definitions in their reporting. It is important to focus on the source used in the resolution criteria.

(10) Pre-mortem. Think about how you are most likely to be wrong. Imagine that you are writing a letter to your future self that you will open and read once the outcome is known. In the letter you try to explain to your future self the most likely way that your forecast will be deemed to be a poor forecast. Are you most worried about missing a key piece of information? What is the biggest uncertainty and would keep you awake at night?

(11) More general advice:
    - Even if something seems impossible, think twice before forecasting less than 3%. (It is possible that you don't have all of the information, or have misunderstood something.)
    - Even if something seems certain, think twice before forecasting more than 97%. (It is possible that you don't have all of the information, or have misunderstood something.)
    - Pay close attention to the exact wording and resolution source in the resolution criteria. Sometimes newspaper articles will cite a number that is significantly different from the number in the resolution criteria. Make sure to pay attention to the resolution criteria.
    - Like a good forecaster, you should use your own judgment to come to the most accurate forecast!

Before answering, as part of your reasoning, you write:
(a) The time left until the outcome to the question is known.
(b) The status quo outcome if nothing changed.
(c) Think about answering the question with different scopes to help ensure that you have a self consistent view and have considered the broader context.
    For example, imagine the question is: Will a company declare bankruptcy in the next 3 months? It can be useful to force yourself to forecast the probability of bankruptcy over the next 1 year, 2 years, and 5 years. Doing this in a self-consistent way helps to force you to consider the scope explicitly.
    If you forecast 40% in 12 months, then you might forecast 10% in 3 months to be scope sensitive. However, it is also possible that the bankruptcy risk is higher in the near term, so it could still be 25%.
    You will need to use your judgment. Being scope insensitive is a common cognitive bias and this exercise is meant to help combat this bias by forcing you to explicitly consider the question's scope.
(d) A brief description of a scenario that results in a No outcome.
(e) A brief description of a scenario that results in a Yes outcome.

The last thing you write is your final answer as a probability between 0 and 1.
\end{lstlisting}
\end{tcolorbox}

\subsection{Resolution}
\label{sec:appendix-resolution}

\begin{figure}[h]
  \centering
  \includegraphics[width=0.99\textwidth]{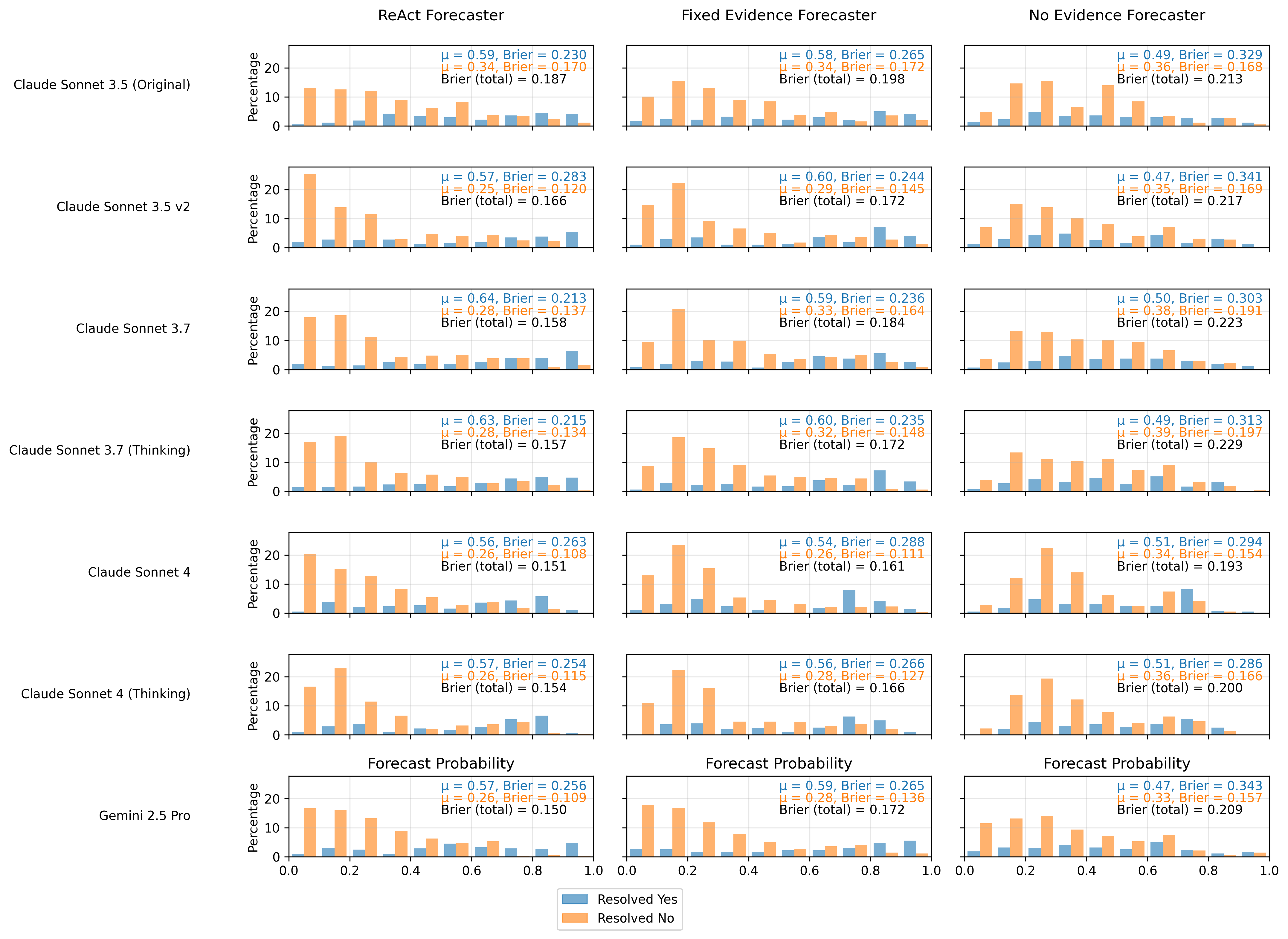}
  \caption{Visualization of resolution for each model and forecasting approach. Frequencies for the bar charts are weighted by the question weight and normalized to sum to 100\% (i.e. questions that resolved Yes and No together sum to 100\%).}
  \label{fig:resolution-benchmark}
\end{figure}

\end{document}